\documentclass[10pt,twocolumn]{article} 
\usepackage{times}
\usepackage{graphicx}
\usepackage{amssymb}
\usepackage{url,hyperref}
\usepackage{color,soul}
\usepackage{adjustbox}
\usepackage{makecell}
\makeatletter
\newcommand{\printfnsymbol}[1]{%
  \textsuperscript{\@fnsymbol{#1}}%
}
\makeatother

\date{}

\begin{document}

\title{\vspace{-1.5em}Efficient large-scale image retrieval with deep feature orthogonality and Hybrid-Swin-Transformers \vspace{1em}}
\author{
Christof Henkel \\
 NVIDIA\\
  \texttt{chenkel@nvidia.com} \\

}

\maketitle

\begin{abstract}

We present an efficient end-to-end pipeline for large-scale landmark recognition and retrieval. We show how to combine and enhance concepts from recent research in image retrieval and introduce two architectures especially suited for large-scale landmark identification. A model with deep orthogonal fusion of local and global features (DOLG) using an EfficientNet backbone as well as a novel Hybrid-Swin-Transformer is discussed and details how to train both architectures efficiently using a step-wise approach and a sub-center arcface loss with dynamic margins are provided. Furthermore, we elaborate a novel discriminative re-ranking methodology for image retrieval. The superiority of our approach was demonstrated by winning the recognition and retrieval track of the Google Landmark Competition 2021.

\end{abstract}

\section{Introduction}

The Google Landmark Dataset v2 (GLDv2) has become a popular dataset for evaluating performance of architectures and methods used for solving large-scale instance-level recognition tasks \cite{weyand2020google}. The original dataset consists of over five million images with over 200,000 classes, originating from local experts who uploaded to Wikimedia Commons. Besides its size, the dataset poses several interesting challenges such as long-tailed distribution of classes, intra-class variability and noisy labels. Since 2019, GLDv2 has been used to asses and test state-of-the art instance-level recognition methods as part of the Google Landmark Competition hosted on kaggle.com. The winning solution of 2019 led to a cleaned version of GLDv2, which is a subset with 1.5 million images containing 81,313 classes and will be denoted by GLDv2c in the following. Furthermore, GLDv2x will be used to denote the subset of GLDv2 which is restricted to the 81,313 classes present in GLDv2c but was not cleaned. The yearly competition is divided into a recognition and retrieval task. The recognition track is about correctly classifying a landmark for a set of test  images, where a significant amount of non-landmarks used as distractors are present. It is evaluated using Global Average Precision (GAP) \cite{perronnin2009family, weyand2020google} as metric. For retrieval the task is to find similar images in a database with respect to a given query image and is evaluated using mean Average Precision@100 (mAP). In contrast to recent years the 2021 competition is evaluated with special focus on a more representative test set\footnote{see \cite{kim2021towards} for details}. The competition follows a code competition setup, where participants are asked to upload their solution code rather than raw predictions. The submitted code will infer on a hidden test set offline, where resources available in the offsite environment are restricted to 12h runtime using a two-core CPU and a P100 GPU. During the competition participants are evaluated on the test set and a part of the predictions is used to calculate the above mentioned metrics and show the best score for each participant on a public leaderboard. After the competition the score with respect to the remaining part is released and used to determine and display the final scoring (private leaderboard). For training our models we used pytorch with mixed precision using 8xV100 NVIDIA GPUs with distributed data parallel (DDP). Moreover, we use several implementations and pretrained weights from timm \cite{rw2019timm}. Our training code will be available online\footnote{\fontsize{7}{9}\selectfont{https://github.com/ChristofHenkel/kaggle-landmark-2021-1st-place}}.

\section{Methodology}

\subsection{Validation strategy}

For recognition track we use the same local validation as in the last years winners solution \cite{henkel2020supporting}, which leverages the 2019 test set and respective post-competition released test labels. 
The retrieval task is assessed in a similar way using the 2019 test query and index dataset together with the post-competition released retrieval solution. However, in our local validation we only considered index images that are matches of any query image to significantly reduce the computation time for evaluation. With this approach we achieved a very good correlation between local validation and leaderboard. For both tasks we tracked the respective competition metric during experiments at the end of every training epoch.

\subsection{Modeling}
\label{sec:modeling}

For both tracks we developed deep learning based architectures, that learn an image descriptor, a high dimensional feature vector, which enables to differentiate between a larger number of classes yet allows for intra-class variability. Although historically global- and local landmark descriptors are trained separately and predictions are combined in a two-stage fashion, attempts are made to not only include the training of local descriptors in a single architecture (e.g. \cite{noh2017large}, \cite{cao2020unifying}) but also omit spatial verification and fuse global and local descriptors within a single-stage model (see \cite{yang2021dolg}). Given a tight competition timeline and restricted inference run-time we focused on single-stage models resulting in a single image descriptor. However, our modeling efforts put local features in the focus as they are especially important for landmark identification.

In the following we present two architectures especially suited for large-scale image recognition/ retrieval with noisy data and high intra-class variability. 
Both conceptually share a large part of an EfficientNet \cite{tan2019efficientnet}  based convolutional neural network (CNN) encoder and a sub-center arcface classification head with dynamic margins \cite{deng2020sub}, which was shown to be superior to classical arcface as demonstrated by the 3rd place 2020 recognition solution \cite{ha2020google}. For training we resize all image to a square size and apply shift, scale, rotate and cutout augmentation using albumentations \cite{albu_2020}. We
use Adam optimizer with weight decay and learning rate and batch size varying per model. We follow a cosine annealing learning rate schedule with one warm-up epoch.

\begin{figure*}[h!]
  \begin{center}
    \includegraphics[width=4.5in]{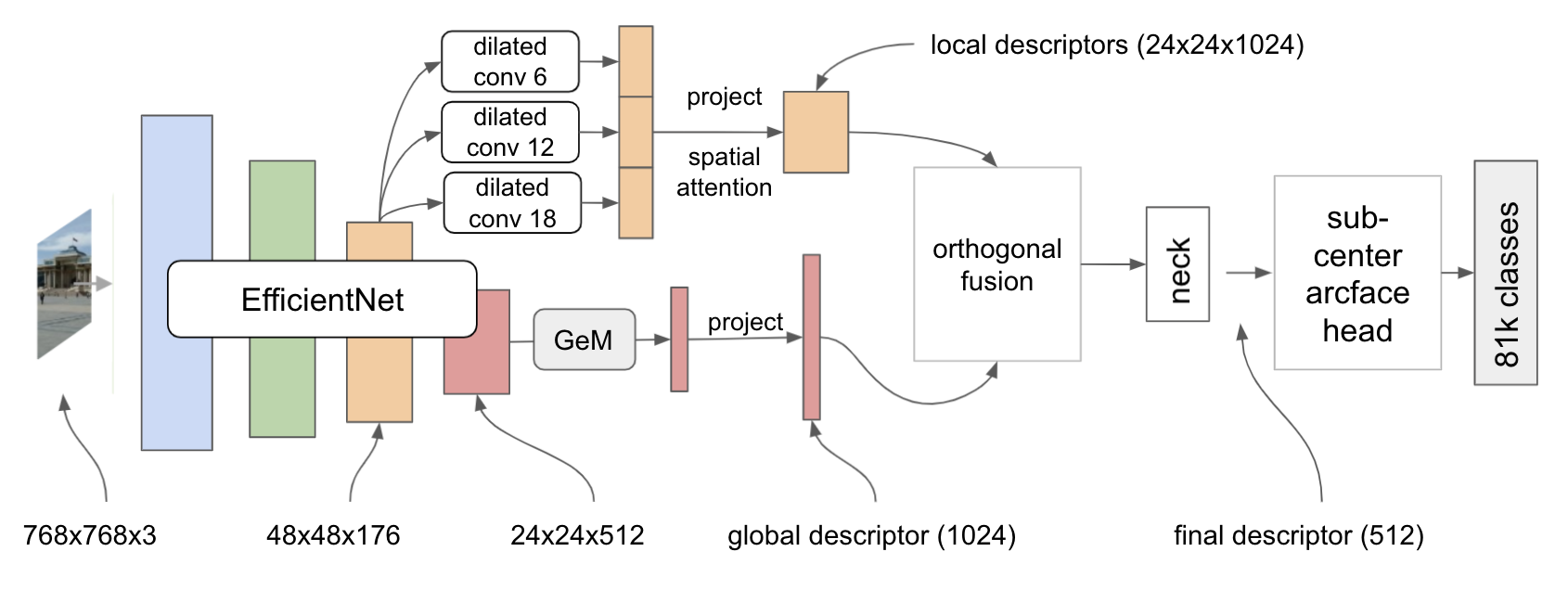}
  \end{center}
  \vspace*{-5mm}
\caption{\small Model architecture DOLG-EfficientNet-B5}
  \label{dolg_arch}
\end{figure*}

\subsubsection{DOLG-EfficientNet with sub-center arcface}
We implemented DOLG \cite{yang2021dolg}, but with some adjustments to improve the performance. Firstly, we used an EfficientNet encoder, which was pretrained on ImageNet. We added the local branch after the third EfficientNet block and extract 1024-dimensional local features using three different dilated convolutions, where dilation parameters differ per model. The local features of the three dilated convolutions are concatenated in feature dimension and self-attended using spatial-2d-attention. The local features are then fused orthogonally with the global feature vector, which resulted from a GeM pooling \cite{radenovic2018fine} of the fourth EfficientNet block output projected to 1024 dimensions.\footnote{see \cite{yang2021dolg} for details of local branch and orthogonal fusion module} The fused features are aggregated using average pooling before they are fed into a neck consisting of a fully connected layer followed by batch-norm and parameterized ReLU activation (FC-BN-PReLU)\footnote{see neck Option-D from \cite{deng2019arcface} for detailed description and rational} resulting in a 512-dimensional single descriptor. For training, this single descriptor is fed into a sub-center (k=3) arcface head with dynamic margins predicting 81,313 classes.

Our DOLG-EfficientNet models are trained following a 3-step procedure. Firstly, the models are trained for ten epochs on GLDv2c using a small image size. Then training is continued for 30-40 epochs on the more noisy GLDv2x using a medium image size. Finally, the models are finetuned for a few epochs on a large image size also using GLDv2x.

\subsubsection{Hybrid-Swin-Transformer with sub-center arcface}

The second architecture leverages recent advances in using transformers for computer vision problems. We appended a vision transformer to a CNN-encoder resulting in a Hybrid-CNN-Transformer model. As such the CNN-part can be interpreted as a local feature extractor, while the transformer-part acts as a graph neural net on those local features aggregating them to a single vector. More precisely we used a Swin-Transformer\cite{liu2021swin} as it  especially flexible at various scales. As input for the transformer we use the output of a few blocks of EfficientNet which is flatten, position embedded and projected to match the transformer dimensions and extended with a virtual token representing the aggregated information. After passing through the Swin-Transformer we fed the features of the virtual token into a FC-BN-PReLU neck to derive the final descriptor. The hybrid model is trained using an sub-center arcface head with 81,313 classes and dynamic margins. 

When combining an EfficientNet encoder and a Swin-Transformer which both have been pre-trained on ImageNet individually, a careful training recipe is important to avoid overflow and other training issues resulting in NaNs especially when using mixed precision. We used the following four-step approach. Firstly, initialize the transformer-neck-head part by training on a small image size using GLDv2c for 10 epochs. Next, exchange the transformers orignal patch embedding with a 2-block EfficientNet encoder, freeze the previously transformer-neck-head part and train the newly added CNN part for one epoch on medium size images. thirdly, unfreeze and train the whole model for 30-40 epochs using GLDv2x. Finally, insert a further pretrained EfficientNet block between the CNN and Swin-Transformer and finetune the whole model for a few epochs using large images and GLDv2x.

\begin{figure*}[h!]
  \begin{center}
    \includegraphics[width=4.5in]{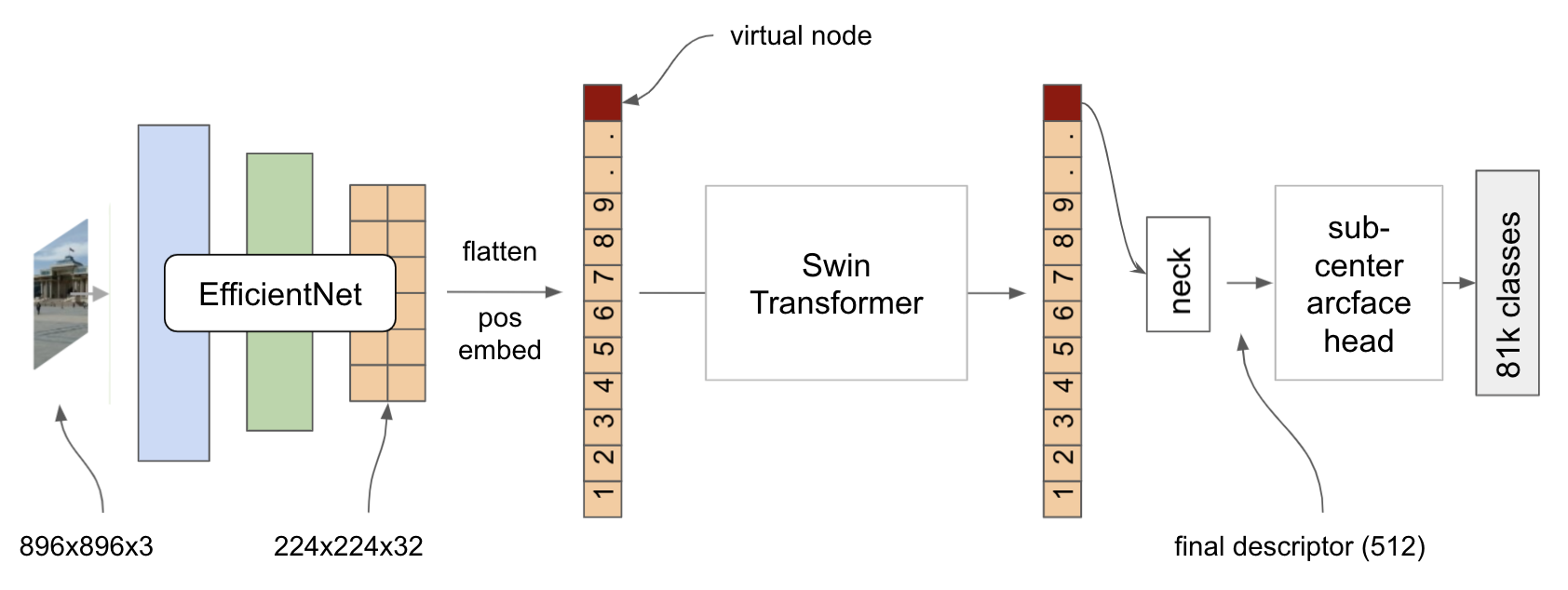}
  \end{center}
  \vspace*{-5mm}
\caption{\small Hybrid-Swin-Transformer, exemplary shown for EfficientNet-B5-Swin-Base224}
  \label{swin_arch}
\end{figure*}

\subsection{Submission ensemble}

The winning submission for recognition track is an ensemble of eight models including three DOLG and three Hybrid-Swin-Transformer models with varying EfficientNet backbones and input image sizes. We additionally trained two pure EfficientNet models  following code and instructions provided by the 3rd place team of google landmark recognition competition 2020 \cite{ha2020google} where models are trained on the full GLDv2. 
For the winning submission of the retrieval track we used nearly the same ensemble but exchanged one of the pure EfficientNet models with an EfficientNet trained following the procedure of 2nd place team of google landmark recognition competition 2020 \cite{bestfitting2020}. Table \ref{model_ensemble_overview} gives an overview of backbones, image sizes, data used and resulting scores. Instead of increasing the image size for the last training step, we reduced the stride in the first convolutional layer from (2,2) to (1,1) for some models, being especially beneficial for small original images. 
\begin{table}[h]
\centering
\begin{adjustbox}{width=\columnwidth,center}
\begin{tabular}{|c|c|c|c|c|c|c|c|}
  \hline
    model & \makecell{image\\size} & stride & data & \makecell{private score\\recognition} & \makecell{public score\\recognition} & \makecell{private score\\retrieval} & \makecell{public score\\retrieval}  \\
    \hline
    DOLG-EfficientNet-B5 & 768 & 2 & GLDv2x & 0.476 & 0.497 & 0.478 & 0.464 \\
    DOLG-EfficientNet-B6 & 768 & 2 & GLDv2x & 0.476 & 0.479 & 0.474 & 0.454 \\
    DOLG-EfficientNet-B7 & 448 & 1 & GLDv2x & 0.465 & 0.484 & 0.470 & 0.458\\
    EfficientNet-B3-Swin-Base-224 & 896 & 2 & GLDv2x & 0.462 & 0.487 & 0.481 & 0.454\\
    EfficientNet-B5-Swin-Base-224 & 448 & 1 & GLDv2x & 0.462 & 0.482 & 0.476 & 0.443\\
    EfficientNet-B6-Swin-Base-384 & 384 & 1 & GLDv2x & 0.467 & 0.492 & 0.487 & 0.462\\
    EfficientNet-B3 & 768 & 2 & GLDv2 & 0.463 & 0.487 & &\\
    EfficientNet-B6 & 512 & 2 & GLDv2 & 0.470 & 0.484 & 0.454 & 0.441\\
    EfficientNet-B5 & 704 & 2 & GLDv2x &   &   & 0.459 & 0.428\\
    Ensemble Recognition &   &   &   &   0.513 & 0.534 & &\\
    Ensemble Retrieval &   &   &   &   &  & 0.537 & 0.518\\
    
    \hline
  \end{tabular}
   \end{adjustbox}
  \caption{Overview of model ensemble}
  \label{model_ensemble_overview}
 
\end{table}

\subsection{Inference}

For both tracks we used an retrieval approach for prediction, where for each query image most similar reference images are searched in database of index images using cosine similarity between L2-normalized image descriptors. For the recognition task the train set is used as index image database and the landmark label of the most similar train images are used as prediction for a given test image. In contrast, for retrieval an additional database of index images to retrieve most similar images from is pre-defined. For recognition track, to add more intra-class variety, we expand the offsite train set with full GLDv2 and landmark images from WIT \cite{srinivasan2021wit}, both filtered to contain only landmarks of the offline train set. For retrieval we further extend with the index set of the 2019 retrieval competition.

\subsubsection{Retrieval post-processing}

In order to retrieve most similar index images given a query image using our ensemble we rank all index images using cosine similarity of the 512-dimensional descriptor for each model individually, resulting in query-index pair scores. We then re-rank the index images by a discriminative re-ranking procedure derived from the one introduced in the retrieval task of the 2019 Google Landmark competition by the winning team \cite{ozaki2019largescale}, where in a first step we label the query and index images using their top3 cosine similarity to a given train set. However, in contrast to the hard re-ranking procedure illustrated in \cite{ozaki2019largescale}, we used a soft up-rank procedure by adding the top3 index-train cosine similarity to the query-index scores if labels of query and index image match. We saw further benefit when additionally performing a soft down-ranking procedure. We implemented the down-ranking by substracting 0.1 times the top3 index-train cosine similarity if labels of query and index image do not match. For each model in our ensemble we extract the top750 index image ids and related scores for each query image using the above method and aggregate the resulting 6000 scores by summing per each image id before we take the top100 as a final prediction. 

\subsubsection{Recognition post-processing}
We use our ensemble to extract eight 512-dimensional vectors for each train and test image. Vectors are then averaged per model type (DOLG-EfficientNet, Hybrid-Swin-Transformer, pure EfficientNet) resulting in three 512-dimensional vectors which are then concatenated leading to a 1536-dimensional image descriptor. We use cosine similarity to find the closest training images for each test image and apply the post-processing from \cite{henkel2020supporting} for re-ranking and non-landmark identification, which results in the final predictions.

\section{Conclusion}

We presented several improvements to previous approaches for large-scale landmark identification leading to winning both tracks of the 2021 Google landmark competition. We showed how deep orthogonal features or vision transformers help to efficiently leverage local information extracted with a CNN backbone and stressed the superiority of sub-center arcface when confronted with long-tailed class distributions and intra-class variability. We confirmed the applicability of the re-ranking and non-landmark identification of \cite{henkel2020supporting} to the more balanced 2021 test set and explained a novel soft discriminative up- and down-ranking procedure for the retrieval task.

\newpage
\bibliographystyle{abbrv}
\bibliography{refs}
\end{document}